\documentclass[10pt,twocolumn,letterpaper]{article}

\usepackage[pagenumbers]{iccv} 

\usepackage{xspace}

\newcommand{\myparagraph}[1]{\noindent \textbf{#1:}}

\def\eg{\textit{e.g.}\@\xspace} 
\def\ie{\textit{i.e.}\@\xspace} 

\newcommand{\abbrv}[0]{\mbox{EditCLIP}\@\xspace}
\newcommand{\ect}[0]{\mbox{\textbf{\texttt{EC2T}}}\@\xspace}
\newcommand{\ecec}[0]{\mbox{\textbf{\texttt{EC2EC}}}\@\xspace}

\AtEndPreamble{
    \usepackage[capitalize]{cleveref}
    \crefname{section}{Sec.}{Secs.}
    \Crefname{section}{Section}{Sections}
    \Crefname{table}{Table}{Tables}
    \crefname{table}{Tab.}{Tabs.}
}

\definecolor{iccvblue}{rgb}{0.21,0.49,0.74}
\usepackage[pagebackref,breaklinks,colorlinks,allcolors=iccvblue]{hyperref}
\usepackage{comment}
\usepackage{colortbl}

\def\paperID{6088} 
\def\confName{ICCV}
\def\confYear{2025}

\title{EditCLIP: Representation Learning for Image Editing }

\author{Qian Wang\and
Aleksandar Cvejic\and
Abdelrahman Eldesokey\and
Peter Wonka\and
KAUST, Saudi Arabia\\
{\tt\small first.last@kaust.edu.sa}
}

\begin{document}
\twocolumn[{%
\renewcommand\twocolumn[1][]{#1}%
\maketitle
\begin{center}
    \centering
    \captionsetup{type=figure}
    \includegraphics[width=0.95\textwidth,trim={0cm 7cm 11cm 0},clip]{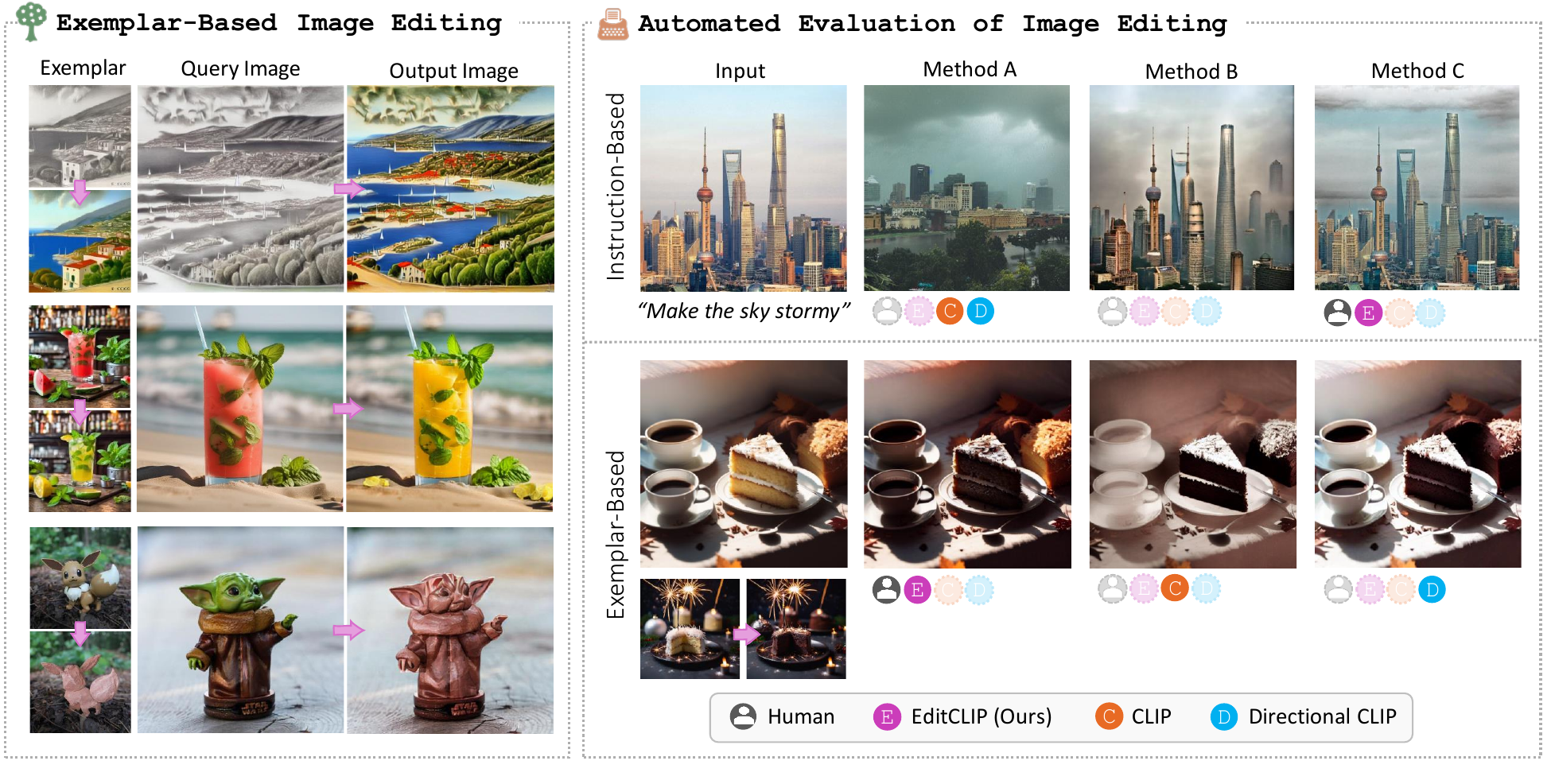}
    \captionof{figure}{\abbrv provides a unified representation of image edits by encoding the transformation between an image and its edited counterpart within the CLIP space. We demonstrate the effectiveness of \abbrv embeddings in exemplar-based image editing and automated evaluation of image editing pipelines, where it achieves better alignment with human assessment.
}
    \label{fig:teaser}
\end{center}%
}]
\begin{abstract}
We introduce \abbrv, a novel representation-learning approach for image editing.  
Our method learns a unified representation of edits by jointly encoding an input image and its edited counterpart, effectively capturing their transformation.  
To evaluate its effectiveness, we employ \abbrv to solve two tasks: exemplar-based image editing and automated edit evaluation.  
In exemplar-based image editing, we replace text-based instructions in InstructPix2Pix \cite{brooks2023instructpix2pix} with \abbrv embeddings computed from a reference exemplar image pair.  
Experiments demonstrate that our approach outperforms state-of-the-art methods while being more efficient and versatile.  
For automated evaluation, \abbrv assesses image edits by measuring the similarity between the \abbrv embedding of a given image pair and either a textual editing instruction or the \abbrv embedding of another reference image pair.  
Experiments show that \abbrv aligns more closely with human judgments than existing CLIP-based metrics, providing a reliable measure of edit quality and structural preservation. 
The code and model weights are available at \href{https://github.com/QianWangX/EditCLIP}{https://github.com/QianWangX/EditCLIP}.
\end{abstract}    
\section{Introduction}
\label{sec:intro}

Image editing is a fundamental task in creative domains such as design and digital art, enabling creators to iteratively refine their creations to align with their artistic vision. 
Recent advancements in diffusion models \cite{sd,dalle2,imagen,blackforestlabs2024flux1dev,Esser2024sd3} have revolutionized image editing \cite{masactrl,brack2024ledits,huberman2024edit,Xu_2024_CVPR_INFEDIT,brooks2023instructpix2pix,hertz2022prompt,geng2024motion,shi2024dragdiffusion,kulikov2024flowedit,wu2024turboedit}, leveraging their deep semantic understanding of images and artistic styles to apply highly realistic edits.
Traditionally, diffusion-based editing approaches rely on textual instructions to specify the desired edits. 
Then, the internal dynamics of a diffusion model are manipulated to localize regions of interest and apply the edit.
While effective, instruction-based editing is limited by the diffusion model's understanding of language and the inherent limitations of natural language in describing complex edits, \eg artistic styles with no established name and compound edits.

Several research directions have emerged to tackle these challenges, either by enhancing the semantic understanding of diffusion models to enable more complex and fine-grained edits \cite{brooks2023instructpix2pix,brack2023Sega,cvejic2025partedit,Xu_2024_CVPR_INFEDIT,huberman2024edit,huang2024smartedit,koh2023gill,ge2024seedllama,ge2023seed} or by incorporating visual prompts \cite{nguyen2023visual,zhao2024instructbrush,srivastava2024reedit,lai2024unleashing,yang2023imagebrush} as a conditioning signal for diffusion models to perform exemplar-based editing.
However, these research efforts face two major bottlenecks. 
First, they still rely on text to specify the edits. 
Even when visual exemplars are provided, they are ultimately mapped to a textual space either through Vision-Language Models (VLMs) \cite{lai2024unleashing,srivastava2024reedit} or by optimizing special textual tokens based on the exemplars \cite{nguyen2023visual,yang2023imagebrush}.

Second, the evaluation of these approaches heavily relies on CLIP-based metrics \cite{clip,Kim_2022_CVPR_dirclip}, which either measure the alignment between the edited image and the textual descriptions or compute a directional embedding vector between the original and edited images.  
However, these metrics primarily assess whether the edit is applied, disregarding whether the structure of the edited image deviates significantly from the original.
Due to this limitation, researchers often rely on human evaluations through user studies to assess edit quality, which incurs higher costs and longer evaluation times.

We propose \abbrv, a novel representation-learning approach for image editing that addresses these challenges altogether by learning an implicit representation of edits beyond linguistic constraints. 
Inspired by CLIP’s ability to capture semantic relationships between images and texts, our method models the semantic relationships between image edits and their corresponding editing instructions within the CLIP space. 
Specifically, our model learns a unified representation of edits by encoding how reference images are transformed into their edited counterparts in relation to the provided instruction. 
We demonstrate the effectiveness of our model on two tasks: \emph{exemplar-based image editing} and \emph{automated evaluation of image editing} tasks.

In exemplar-based editing, given a single example of an image and its edited counterpart, our \abbrv embedding is computed and used to guide the diffusion process to replicate the edit on a new output image without requiring textual editing instruction. 
This capability enables complex and precise edits, where describing the edit in natural language is challenging.
For instance, an artist who applies multiple edits to an image but struggles to describe them in words can use \abbrv to capture and transfer the edits seamlessly. 
Experiments show that our approach outperforms existing exemplar-based image editing methods across different types of edits and even outperforms the recent state-of-the-art approach InstaManip \cite{lai2024unleashing} despite having only $5.9\%$ the number of parameters.

For automated evaluation of image editing, we measure the edit-instruction alignment by computing the similarity between the \abbrv embedding of a given image pair and either the embedding of the textual editing instruction or the \abbrv embedding of another reference image pair. 
Unlike CLIP-based metrics that independently embed images and compute differences between their global visual embeddings, \abbrv embeddings directly capture how the image is transformed, taking into consideration how the edit is applied and if the unedited regions are preserved. 
Experiments show that \abbrv aligns more closely with human judgments than existing CLIP-based metrics, providing a scalable and automated alternative for evaluating image editing methods. 
By streamlining evaluation, our approach can help accelerate the research of image editing.


\noindent Our contributions can be summarized as follows:
\begin{itemize}
    \item We propose {\abbrv}, a representation-learning approach that produces a unified representation for various types of image edits.  
    \item We show that the learned representations can be used for exemplar-based image editing, replacing text-based instructions in diffusion models.  
    \item We further show that \abbrv provides a reliable edit representation, enabling the assessment of both edit quality and faithfulness to the reference image.  
\end{itemize}


\section{Related Work}
\label{sec:related}

\subsection{Diffusion-Based Image Editing}
The emergence of image diffusion models has driven the development of powerful image editing approaches, leveraging their deep understanding of image semantics.
One category of these approaches is training-free that either manipulates the internal representations of the diffusion U-Net \cite{masactrl,alaluf2024cross,chung2024style,hertz2022prompt,xu2023infedit,liu2024towards,pnp}, or manipulate the diffusion trajectory \cite{Wu_2023_ICCV,brack2024ledits,huberman2024edit,han2023improving} to achieve the desired edits. 
Another category fine-tunes a pre-trained diffusion model on image editing datasets, enabling it to apply edits \cite{brooks2023instructpix2pix,guo2024focus,zhang2023magicbrush,kawar2023imagic}. 
Alternatively, test-time optimization was employed in \cite{choi2023custom,wang2023dynamic,unitune,dong2023prompt,nam2024contrastive} to perform customized edits given a single image.
In all these approaches, the textual embedding of an editing instruction serves as a condition to steer the diffusion model toward the intended edit.

\begin{figure*}
    \centering
    \includegraphics[width=\linewidth]{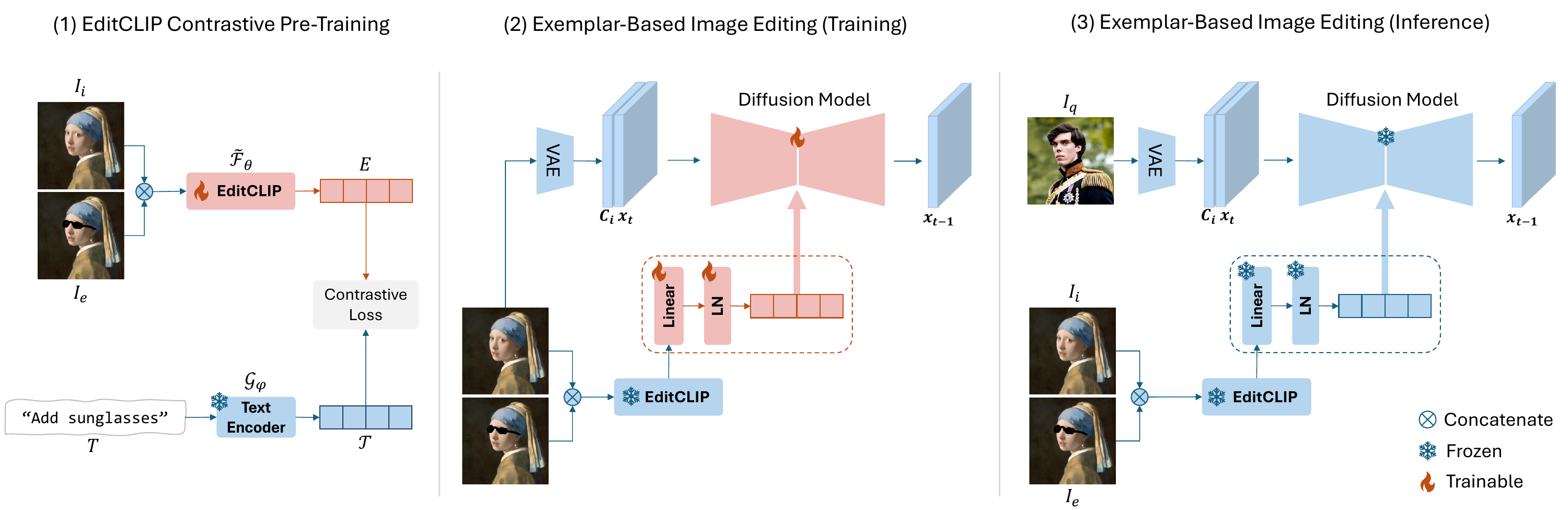}
    \caption{
    An overview of our proposed approach. \abbrv is pre-trained similarly to CLIP, but the visual encoder processes a concatenated exemplar image pair. After pre-training, \abbrv can replace the text encoder in InstructPix2Pix \cite{brooks2023instructpix2pix} to enable exemplar-based editing.}
    \label{fig:method}
\end{figure*}


\subsection{Exemplar-Based Image Editing}
A major limitation of instruction-based editing approaches is their reliance on language to describe the edit, which can be challenging for complex edits, especially when multiple edits are combined.
Exemplar-based image editing addresses this issue by performing edits based on a user-provided reference image pair. 
Prior work \cite{wang2023promptdiffuison} aimed at solving in-context learning tasks but could also handle exemplar-based image editing, by projecting a reference image embeddings into a ControlNet~\cite{zhang2023controlnet}.
Approaches such as \cite{nguyen2023visual,yang2023imagebrush} encoded the edit by optimizing special textual tokens derived from the reference image pair, which can then be used to apply the edit to new query images. 
Nonetheless, these methods are mostly limited to stylistic edits, and they may fail when there is more disparity between the exemplar and the query image, restricting their applicability to more diverse editing tasks.

Other works \cite{srivastava2024reedit,lai2024unleashing} attempted to leverage Vision-Language Models (VLMs) to describe the edit between image pairs. 
Similarly, \cite{zhao2024instructbrush} optimized instruction pairs to represent the transformation between the exemplar pair.
However, these approaches remain constrained by linguistic descriptions and introduce significant computational overhead due to the complexity of VLMs or the need for costly optimizations.
In contrast, our proposed \abbrv produces an implicit representation of edits, making it unconstrained by linguistic descriptions.
This allows it to better capture complex edits that are difficult to express in natural language. 
Moreover, \abbrv serves as a plug-and-play substitute for the CLIP text encoder in diffusion models, making it seamlessly integrated into popular editing pipelines such as InstructPix2Pix \cite{brooks2023instructpix2pix}.


\subsection{Evaluating Image Editing Approaches}
A key aspect when developing image editing approaches is the evaluation protocol. 
A common practice is to use CLIP score \cite{clip} between the image embedding of the edited image and the text embedding of the editing instruction. 
However, this approach does not account for how much the edited image deviates from the original.
To address this, previous works \cite{Kim_2022_CVPR_dirclip,nguyen2023visual} have proposed directional CLIP score (CLIP directional similarity), which compares the directional embedding between the source and edited image with either the editing instruction or a directional embedding from a reference editing pair \cite{nguyen2023visual}.
Nonetheless, these metrics only focus on how the edit is globally applied and do not take into account if the structure of the source image is preserved.
In contrast, our proposed \abbrv explicitly encodes the difference between the source and edited image, \ie, the edit itself, effectively capturing both the edit and the deviation from the source image for a more accurate and detailed evaluation.

\section{Method}
\label{sec:method}

We aim to design a representation learning approach for image editing, where edits can be implicitly encoded within an embedding space. 
Below, we introduce \abbrv, a model designed to learn a general representation of edits.
We first describe our approach and then analyze how our model captures edit semantics. 
Then, we explain how \abbrv can be used for exemplar-based image editing as an alternative to text-based editing instructions.
Finally, we demonstrate the versatility of \abbrv embeddings by employing them for automated evaluation of image editing pipelines.


\subsection{Representation Learning for Image Editing }
In image editing, given an input image ${I_i \in \mathbb{R}^{H \times W \times C}}$, the objective is to produce an edited image ${I_e \in \mathbb{R}^{H \times W \times C}}$ based on a textual instruction $T$. 
This transformation can be formulated as:
\begin{equation}
\label{eq:text_edit}
    I_e = \mathcal{U}(I_i; T)\enspace,
\end{equation}
where $\mathcal{U}$ represents the editing pipeline that modifies $I_i$ according to $T$.
A key challenge lies in determining the level of detail required in the textual instruction $T$ to achieve the intended edit. 
Ideally, $T$ should specify how every element of $I_i$ is transformed into $I_e$, but this is sometimes infeasible. 
Instead, an effective approach should aim to capture the transformation from $I_i$ to $I_e
$ in a more structured and learnable manner.

This problem shares similarities with representation learning of images and text in CLIP \cite{clip}, where the goal was to learn a shared representation of images and text.
CLIP has been shown to effectively capture semantic relationships between images and text from relatively coarse textual descriptions using contrastive learning on large-scale image-text pairs. 
Following this strategy, we aim to learn the semantics of edits within the CLIP space, leveraging its ability to encode meaningful transformations from textual guidance.


\subsection{\abbrv Pre-Training}
\label{sec:pretraining}
A standard CLIP model consists of a visual encoder $\mathcal{F}_\theta$ and a text encoder $\mathcal{G}_\phi$\enspace, parameterized by learnable parameters $\theta$ and $\phi$, respectively. 
Our objective is for the visual encoder to capture how the input image $I_i$ is semantically and visually transformed into the edited image $I_e$. 
To achieve this, we modify the visual encoder $\mathcal{F}_\theta$ to accept a composite input image,
where the input and edited images are concatenated along the channel dimension. 
We denote this new encoder as $\mathcal{\tilde{F}}$ and it produces an edit embedding $E$ as:
\begin{equation}\label{eq:ec}
        E = \mathcal{\tilde{F}}_\theta(\texttt{concat}(I_i, I_e)) \in \mathbb{R}^{d_{e} \times 768},
\end{equation}
where $d_{e}$ is the number of tokens for $E$. 
For the text encoder, we encode the editing instruction $T$ into textual embedding $\mathcal{T}$, which describes the transformation from $I_i$ to $I_e$, as:
\begin{equation}
    \mathcal{T} = \mathcal{G}_\phi(T) \in \mathbb{R}^{d_{t} \times 768} \enspace,
\end{equation}
where $d_{t}$ is the number of tokens for $\mathcal{T}$. Following the contrastive learning paradigm of CLIP, we align the learned \emph{editing space} with the textual space, where we train only the visual encoder while keeping the pre-trained textual encoder frozen.
The training data is sampled from existing instruction-based image editing benchmarks, which provide triplets consisting of an input image $I_i$, its edited counterpart $I_e$, and the corresponding editing instruction $T$.

\begin{figure}
    \centering
    \includegraphics[width=\columnwidth]{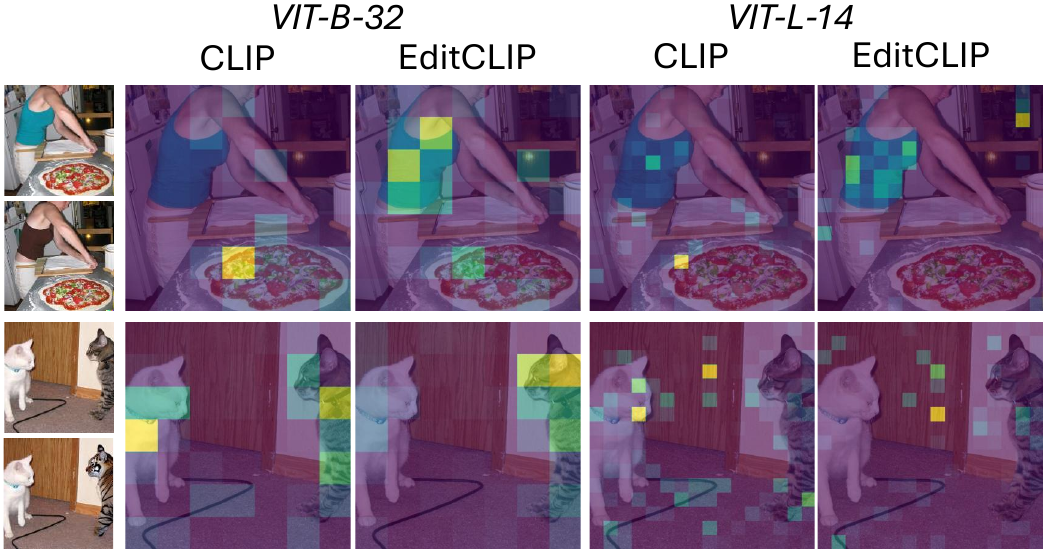}
    \caption{A visualization of the visual encoder’s attention in \abbrv compared to the original CLIP. We visualize the attention of the $[CLS]$ token from the last attention head. Unlike CLIP, where attention is dispersed across the image, \abbrv focuses on the differences between the input and edited image, indicating that it effectively captures the edited regions.}
    \label{fig:attn}
\end{figure}

To analyze what the \abbrv model learns, we follow the common practice of visualizing the attention of the $[CLS]$ token from the last attention head in the final transformer layer of the visual encoder \cite{Chefer_2021_ICCV}. 
As shown in \Cref{fig:attn}, \abbrv focuses on the regions corresponding to the applied edits, such as shifting attention to the woman's torso and the edited cat on the right.


\subsection{\abbrv for Exemplar-Based Image Editing}
To demonstrate the effectiveness of our proposed \abbrv embeddings, we employ them as a substitute for textual editing instructions in Instruct-Pix2Pix (IP2P) \cite{brooks2023instructpix2pix}.  
IP2P is a diffusion-based image editing approach that conditions on a textual editing prompt and an input image to generate an edited output that fulfills the specified edit. 
To train IP2P with our embeddings, we feed the input image $I_i$ and its edited counterpart $I_e$ into \abbrv to obtain the edit embedding $E$, as per \Cref{eq:ec}. 
The same input image is also encoded into the latent space of the diffusion model using the VAE encoder $\mathcal{E}$ that is concatenated with the input noise $x_t$.  

To align $E$ with the text embedding space originally used to train the diffusion model, we process it through a trainable linear layer followed by Layer Normalization.
Note that we use the last hidden state of the \abbrv visual encoder instead of the projected embedding, but we use $E$ for simplicity.
Finally, the diffusion model is fine-tuned using the standard diffusion noise-prediction loss to learn to denoise the latent of the edited image:
\begin{equation}
\mathcal{L}_{\text{noise}} = \mathbb{E}_{\mathcal{E}(I_e),\, \mathcal{E}(I_i),\, E,\,
\epsilon \sim \mathcal{N}(0,1),\, t}
\Bigl[
\|\epsilon - \epsilon_{\theta}(x_t, t, \mathcal{E}(I_i), E)\|_{2}^{2}
\Bigr].
\end{equation}
where $\epsilon$ is the groundtruth noise added to the noisy latent $x_t$.
The training pipeline is illustrated in \Cref{fig:method}.

To further preserve the layout from the input image, we adopt an LPIPS loss~\cite{zhang2018lpips} between the input image and the reconstructed image $I_0^t$ that is computed at denoising timestep $t$ as:
\begin{equation}
    I_0^t = \mathcal{D}((x_t - \sqrt{1 - \bar{\alpha}_t}\epsilon_\theta) / \sqrt{\bar{\alpha}_t}),
\end{equation}
where $\bar{\alpha}_t$ is the coefficient of the DDPM noise scheduler \cite{ho2020denoising}, $\epsilon_\theta$ is the estimated noise, and $\mathcal{D}$ is the VAE decoder.
The total training objective becomes:
\begin{equation}
\label{eq:loss}
    \mathcal{L}_\text{total} = \lambda_{1} \mathcal{L}_\text{noise} + \lambda_{2} \ \text{LPIPS} \left(I_0^t, I_i \right)
\end{equation}
 where $\text{LPIPS}$ is the model used to compute LPIPS loss, and $\lambda_{1}$ and $\lambda_{2}$ are the loss weighing hyperparameters.

During inference, to apply an edit to a new query image $I_q$, the model is conditioned on the \abbrv embedding produced from the exemplar image pair $I_i$ and $I_e$, while the latent representation of the query image is concatenated with the noise $x_t$.
This effectively modifies \Cref{eq:text_edit} to perform exemplar-based image editing, generating an output image $I_o$, which is the edited version of $I_q$:
\begin{equation}
    I_o = \mathcal{U}(I_q; \mathcal{\tilde{F}}_\theta(\texttt{concat}(I_i, I_e)))\enspace,
\end{equation}
where $\mathcal{U}$ is our editing model.

\subsection{\abbrv for Evaluating Edits }
\label{sec:metrics}
As demonstrated in the \Cref{fig:attn}, \abbrv effectively captures semantic changes between the reference image and its edited counterpart. 
At the same time, the \abbrv embeddings $E$ are trained to exhibit high similarity with the textual embedding $\mathcal{T}$ of the respective editing instruction $T$. 
Leveraging these properties, we can assess how well a performed edit aligns with a given editing textual instruction.

Given an input image $I_i$, an arbitrary editing approach generates an edited image $I_e$ based on the editing instruction $T$.
We define the \abbrv-to-Text (\ect) similarity metric as:
\begin{equation}\label{eq:ec2t}
    \text{\ect} (I_i, I_e, T) = \texttt{cos}(\mathcal{\tilde{F}}_\theta(\texttt{concat}(I_i, I_e)), T) \enspace ,
\end{equation}
where $\texttt{cos}$ denotes cosine similarity.
This metric quantifies how the input image transforms into the edited image and whether the changes align with the specified editing instructions. 
Unlike existing metrics based on the original CLIP, our edit embeddings implicitly capture all changes between the reference and edited images.
This enables the evaluation of complex edits while penalizing undesired changes in the image that were not specified in the editing instructions.

In exemplar-based image editing, where both the reference input image $I_i$ and its edited counterpart $I_e$ are provided, the goal is to apply the same edit to a query image $I_q$ without requiring textual instruction. 
Given an output image $I_o$ produced by an arbitrary exemplar-based editing approach, another metric, \abbrv-to-\abbrv (\ecec), can be computed as:
\begin{equation}\label{eq:ecec}
\begin{split}
    \text{\ecec} (I_i, I_e, I_q, I_o) = \texttt{cos}(&\mathcal{\tilde{F}}_\theta(\texttt{concat}(I_i, I_e)) \\ 
    &\mathcal{\tilde{F}}_\theta(\texttt{concat}(I_q, I_o))) \enspace .
\end{split}    
\end{equation}
This metric would capture how similar the edit is between the reference and the target pairs.




\section{Experiments}
\label{sec:exp}

We demonstrate the effectiveness of our \abbrv model on two tasks: (1) exemplar-based image editing and (2) automated evaluation of image editing methods.  
To ensure reliable evaluation, we complement our experiments with user studies conducted by humans to validate our findings.

\begin{table*}
\centering
\label{table:quantitative_exemplar}
\small
\begin{tabular}{l|c|ccc|cc|cc|c}
\toprule
& & \multicolumn{3}{c|}{\emph{Text-based}} & \multicolumn{2}{c|}{\emph{Exemplar-based}} & \multicolumn{2}{c|}{\emph{User-Study}} & \\
 & LPIPS $\downarrow$ & CLIP $\uparrow$ & \text{\ect} $\uparrow$ & CLIP-Dir. $\uparrow$ & $S_\text{visual}$\cite{nguyen2023visual} $\downarrow$ & \text{\ecec} $\uparrow$ & WR-Edit $\uparrow$ & WR-Pres $\uparrow$ & RT (s) \\
\midrule
IP2P\cite{brooks2023instructpix2pix}*      & {0.295} & {0.226} & {0.196} & {0.232} & {0.710} & {0.418} & 55.64 & 52.31 & 1.8 \\
\midrule
VISII\cite{nguyen2023visual}                & 0.518          & 0.203          & 0.152          & 0.096          & 0.832          & 0.313          & 79.84 & 79.43 & 370 \\
PromptDiffusion\cite{wang2023promptdiffuison} & 0.620          & 0.180          & 0.127          & 0.041          & 0.906          & 0.204          & 91.80 & 89.65 & 5.5 \\
InstaManip\cite{lai2024unleashing}          & 0.383          & \textbf{0.226} & \underline{0.168}          & \textbf{0.161} & \textbf{0.735} & \underline{0.383}          & 51.21 & 53.43 & 14.9 \\

\textbf{EditCLIP (Ours)}                    & \textbf{0.233}  & \underline{0.216} & \textbf{0.180} & \underline{0.143}          & \underline{0.761}          & \textbf{0.477}  & -     & -     & \textbf{1.8} \\
\bottomrule
\end{tabular}
\caption{Quantitative results for exemplar-based image editing. *IP2P is text-based, but we include it as a reference. WR-Edit and WR-Pres denote the winning rate of edit quality and input preservation of \emph{our method against other methods} according to human evaluators. RT refers to runtime in seconds. We show the best one in \textbf{bold font} and second best in \underline{underline}. Our approach performs on par with the recent SOTA method, InstaManip, despite having only 20 times fewer parameters.}
\end{table*}


\subsection{Experimental Setup}

\myparagraph{Training Dataset}
We employ the Instruct-Pix2Pix (IP2P)~\cite{brooks2023instructpix2pix} image editing dataset for both \abbrv pre-training and for exemplar-based image editing.
The dataset is instruction-based and contains around 313k filtered input/edit/instruction triplets \footnote{https://huggingface.co/datasets/timbrooks/instructpix2pix-clip-filtered}. 
The edit types in the dataset primarily consist of global style transfer and local object addition or replacement.


\myparagraph{\abbrv Pre-training} 
We initialize our model from pre-trained CLIP models \cite{clip}, modifying and fine-tuning the visual encoder as explained in \Cref{sec:pretraining} while keeping the text encoder frozen.
We apply a learning rate of $2e-4$ to the first convolution layer, which processes both the reference input and edited image and we use a lower learning rate of $2e-6$ for all other layers.
We experiment with two CLIP variations that are commonly used, \texttt{ViT-B/32} and \texttt{ViT-L/14}.  
Each model is trained until convergence, with the former converging after 35 epochs and the latter after 40 epochs.  
All training was conducted on $4$ NVIDIA A100-80G GPUs with a per-GPU batch size of $256$.  

\myparagraph{Exemplar-based Editing Training and Inference} 
We adopt the base training setup from IP2P, using Stable Diffusion 1.5 \cite{sd} as the base model, and initialize it with the weights from the pre-trained IP2P.
For the loss in \Cref{eq:loss}, we set the weights $\lambda_{1} = 1$ and $\lambda_{2} = 0.05$. 
We use a constant learning rate of $5e-5$ throughout the training and train for $16k$ iterations. 
The training was done on a single NVIDIA A100-80G with batch size 64. 
During inference, we use a fixed edit guidance scale $s_E=7$ for edit embedding $E$ and image guidance scale $s_I=1.5$ for input image $I_i$ (see the supplementary materials for more details).
\begin{figure*}[t!]
\centering
\includegraphics[width=0.95\textwidth]{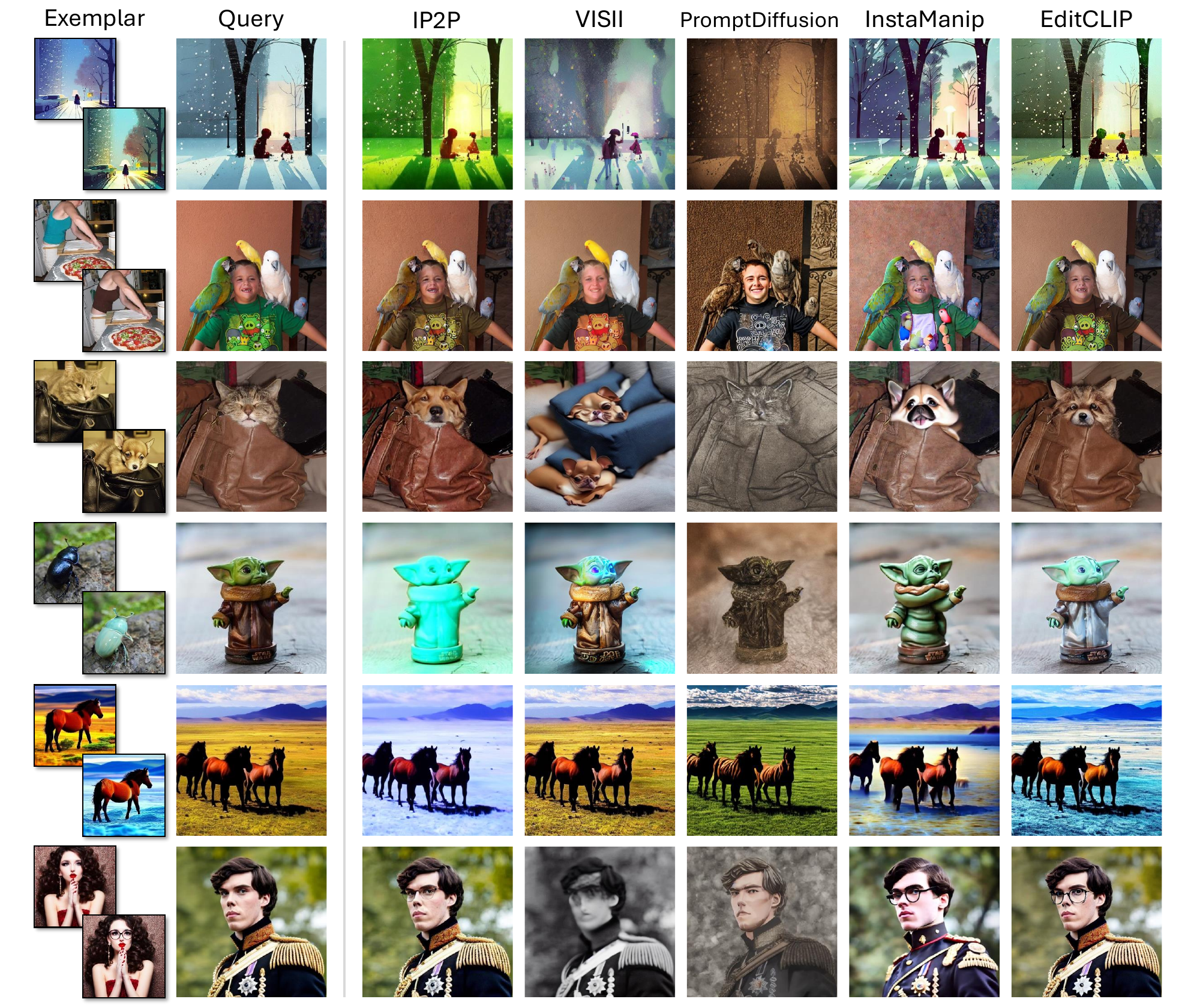}
\caption{Qualitative comparison for exemplar-based image editing.}
\label{fig:qualitative_comp}
\end{figure*}

\myparagraph{Evaluation Benchmark}
We adapt the TOP-Bench dataset \cite{zhao2024instructbrush} for exemplar-based image editing and we denote it as \emph{TOP-Bench-X}.  
TOP-Bench consists of different types of edits, where each type includes a set of training and test pairs.  
We use the training set to form exemplar pairs, denoted as $[I_i, I_e]$, while the test set provides the corresponding query image $I_q$.
This results in a total of 1277 samples, comprising 257 unique exemplars and 124 unique queries.  
We employ this benchmark to evaluate both exemplar-based image editing and the alignment of our proposed metrics with human judgment.  
To assess the perceptual quality of edits, we conducted a two-alternative forced-choice (2AFC) user study on Amazon Mechanical Turk.  
Participants rated two criteria: (1) the quality of the edits and (2) the preservation of query image details (see supplementary materials for further details).

\begin{figure*}[t!]
\centering
\includegraphics[width=0.9\textwidth, trim={0cm 0cm 0cm 0},clip]{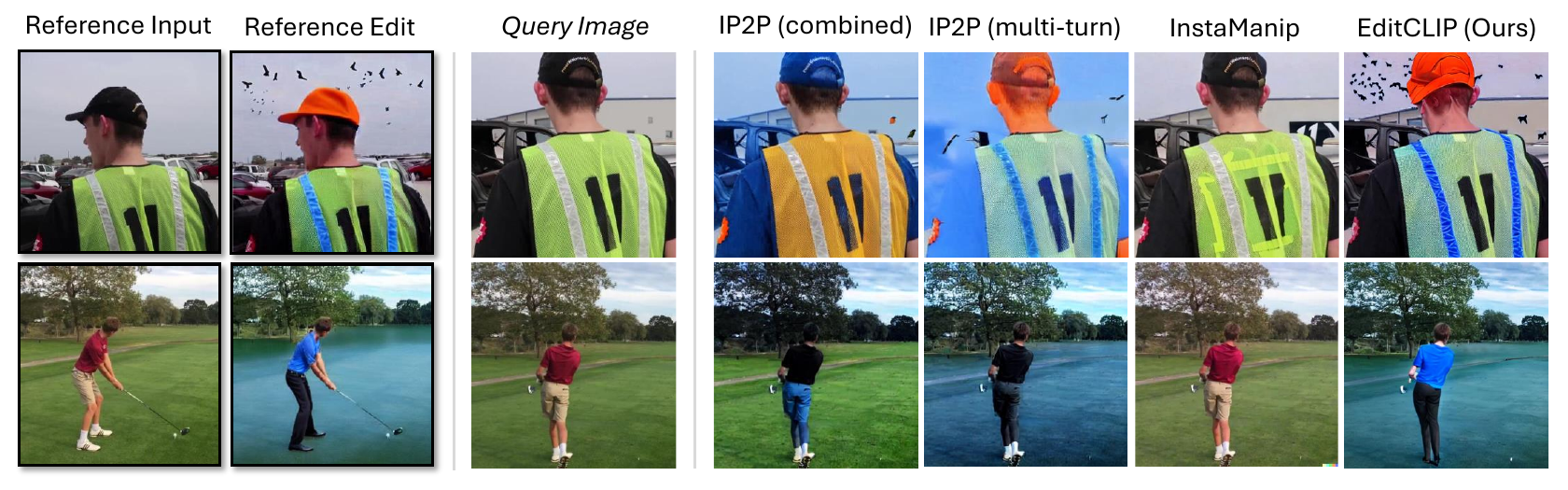}
\caption{\abbrv can perform complex edits when the exemplars contain multiple edits in a single step.}
\label{fig:multiedit}
\end{figure*}

\subsection{Exemplar-based Image Editing}
Here, we demonstrate the capabilities of our proposed \abbrv embeddings as an editing conditioning signal, replacing text-based instructions in image editing.

\myparagraph{Baselines}
We compare against existing exemplar-based approaches with publicly available source code, including 
VISII~\cite{nguyen2023visual}, PromptDiffusion (PD)~\cite{wang2023promptdiffuison}, and the recent InstaManip~\cite{lai2024unleashing}.  
Additionally, we include IP2P \cite{brooks2023instructpix2pix} as a reference for how an instruction-based approach would perform in comparison.
For all methods in comparison, we follow the original setups of their respective code bases. 
For improved credibility, we run each evaluation sample with $5$ different random seeds for every method.


\myparagraph{Quantitative Results} 
\label{para:quant}
We evaluate on \emph{TOP-Bench-X} and report the results in \Cref{table:quantitative_exemplar}.  
We include the \emph{exemplar-based} metric $S_\text{visual}$ \cite{nguyen2023visual}, along with our proposed \ecec metric, described in \Cref{sec:metrics}, and a user study to validate our findings.  
Our approach performs the best on \ecec, a result that is confirmed by the user study with our winning rate larger than $50\%$ against all baselines, demonstrating superior edit quality and better preservation of the query image structure.
We also include the commonly used LPIPS and text-based metrics, including CLIP Score, CLIP Directional Similarity, and our proposed \ect metric for completeness.  
Note that these metrics are computed using the textual editing instruction $T$ provided by the benchmark or textual description of the output image.
As expected, IP2P achieves the best performance on all text-based metrics, as it employs the textual instruction as a conditioning signal.  
Our approach performs the best on \ect among exemplar-based approaches, which aligns with the user study.
In terms of runtime, our method is the fastest, as it neither requires test-time optimization like VISII nor employs large Vision-Language Models (VLMs) as in InstaManip.  
More details on the metrics and the user study can be found in the supplementary materials.


\myparagraph{Qualitative Results}
To facilitate qualitative comparisons with the baselines, we evaluate on selected samples in prior work \cite{wang2024mdp, cheng2024zest, zhang2023magicbrush, srivastava2024reedit, nguyen2023visual, ge2024seeddataedit}.
\Cref{fig:qualitative_comp} shows the qualitative comparison between our method using \abbrv with the \texttt{VIT-L-14} backbone. 
We provide the results obtained by the backbone \texttt{VIT-B-32} in the supplementary materials.
Our approach excels across various types of edits, including global style transfer, color modification, object addition and swapping, and material editing.  
IP2P performs well at edits that are easily described in text, \eg, ``adding glasses'' or ``changing a cat to a dog,'' but struggles with edits such as style or material transfer, as these edits are often difficult to express in text.  
This highlights the effectiveness of our \abbrv embeddings in capturing edits that are not easily described through text.

VISII performs reasonably on style transfer but struggles with other types of edits, as its test-time optimization may diverge.  
The recent state-of-the-art method, InstaManip, demonstrates strong performance across various types of edits; however, this comes at a significant computational cost due to its reliance on a huge VLM.  
In contrast, our method outperforms InstaManip in accurately applying fine details with high fidelity while preserving the original layout, all at a drastically lower computational cost.


\begin{figure*}[ht!]
\centering
\includegraphics[width=1\textwidth]{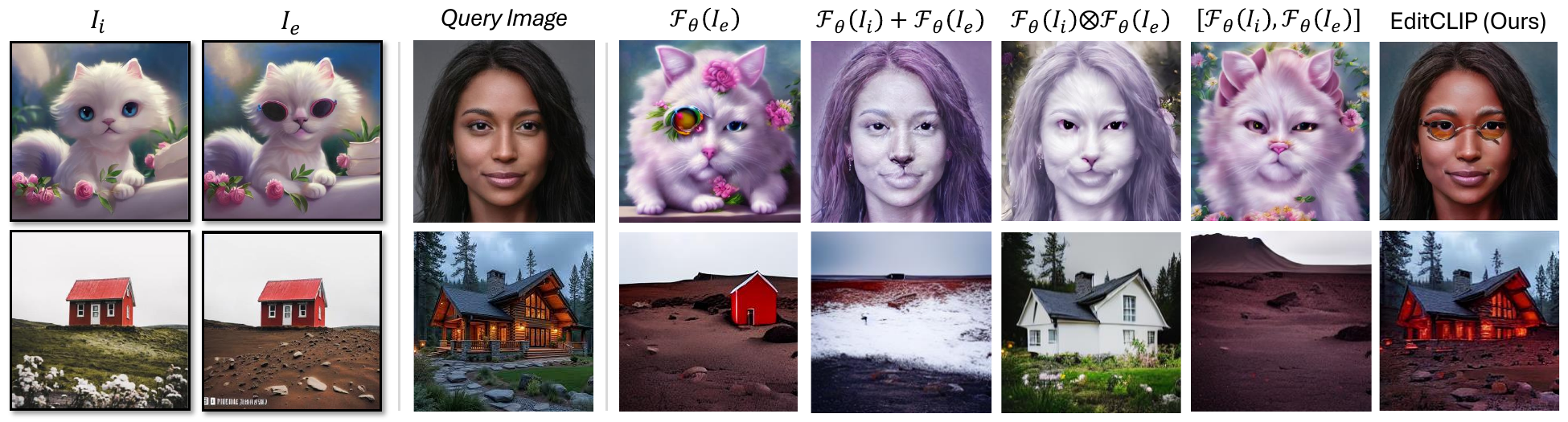}
\caption{Ablation of different conditioning embeddings using the original CLIP visual encoder $\mathcal{F}_\theta$. \abbrv embedding greatly outperforms all CLIP variations.}
\label{fig:ablation_condition}
\end{figure*}

\myparagraph{Multi-Edit Examples} 
To demonstrate the effectiveness of \abbrv in handling complex exemplars with multiple edits, we present challenging editing cases where multiple edits are present in the exemplar.  
For iP2P, we construct two different variations:  
\emph{IP2P (combined)}, which receives a single textual instruction combining all edits and performs them in one step and  
\emph{IP2P (multi-turn)}, which receives separate textual instructions for each edit and applies them sequentially over multiple steps.  
For both InstaManip and our method, all edits are performed in a single shot.  
As shown in \Cref{fig:multiedit}, our method successfully transfers multiple edits from the exemplar in just one shot, while both IP2P and InstaManip fail.


\myparagraph{Ablation Study}
To demonstrate the effectiveness of \abbrv embeddings over the original CLIP, we experiment with different conditioning setups for capturing the edits using the original CLIP.
We only modify the conditioning embedding $E$ to reflect these changes, but we keep all training and inference parameters the same.
The setups that we explore are: 
\begin{itemize}
    \item {$\mathcal{F}_\theta(I_e)$}: Embedding of the reference edit image. 
    \item {$\mathcal{F}_\theta(I_i) + \mathcal{F}_\theta(I_e)$}: Sum of the reference input and edited image embeddings.
    \item {$\mathcal{F}_\theta(I_i) \otimes \mathcal{F}_\theta(I_e)$}: Concatenation of the reference input and edited image embeddings along the channel dim.
    \item {$[\mathcal{F}_\theta(I_i) , \mathcal{F}_\theta(I_e)]$}: Appending the reference input and edited image embeddings along the sequence length dim.  
\end{itemize}  

We present the comparison in \Cref{fig:ablation_condition}.  
$\mathcal{F}_\theta(I_e)$ fails to capture the edit, causing the model to generate only variations of the reference input image.  
For all other variations that utilize both the reference input and edit images, the model struggles to identify the intended edit and instead blends the two images uncontrollably.  
In contrast, our \abbrv embeddings effectively capture the edits and accurately transfer them to the query image without altering its structure.  
Notably, in the first row, the reference exemplar exhibits a slight global style change, making the image more saturated. \abbrv accurately captures this adjustment alongside the intended edit of adding sunglasses.
We provide additional ablation analysis in the supplementary.


\subsection{Automated Evaluation of Image Editing}
To evaluate how well existing CLIP-based metrics, including CLIP, Directional CLIP, and $S_\text{visual}$, as well as our proposed metrics in \Cref{sec:metrics}, align with human evaluation, we compute the Pearson correlation between human judgments and each of these metrics in \Cref{table:table_2_pearson}.
To evaluate \emph{text-based} metrics for instruction-based editing, we processed the TOP-Bench-X benchmark using IP2P and two additional approaches: Ledits++ \cite{brack2024ledits} and EF-DDPM \cite{huberman2024edit}.
For the text-based metrics, our proposed \ect achieves the highest correlation with human judgment both in edit quality and image preservation, indicating a better alignment with humans.
For evaluating \emph{exemplar-based} metrics, we referred to the same user study mentioned in \cref{para:quant}. Our \ecec achieves a higher correlation than $S_\text{visual}$ both in edit quality and image preservation.
These results showcase that our proposed \abbrv embeddings are more reliable metrics for automated evaluation of both instruction-based and exemplar-based image editing methods.

\begin{table}[t!]
\centering
\small
\label{table:table_2_pearson}
\vspace{-2mm}
\resizebox{.47\textwidth}{!}{%
\begin{tabular}{l|ccc|cc}
\toprule
& \multicolumn{3}{c|}{\emph{Text-based}} 
& \multicolumn{2}{c}{\emph{Exemplar-based}} \\
\cmidrule(lr){2-4} \cmidrule(lr){5-6}
 & CLIP
 & CLIP-Dir. 
 & \text{\ect} 
 & $S_{\text{visual}}$ 
 & \text{\ecec} \\
\midrule
Edit & 0.209  & 0.186 & \textbf{0.256} & 0.240 & \textbf{0.372} \\
Preserves & -0.028 & -0.023 & \textbf{0.104} & -0.023 & \textbf{0.157} \\
\bottomrule
\end{tabular}%
}
\caption{Pearson correlation between individual metrics and human judgment in terms of edit quality and input preservation. Our proposed metrics achieve the highest correlation with human evaluation demonstrating better alignment. }
\end{table}

\section{Limitations and Future Work}

\abbrv is trained solely on the IP2P dataset \cite{brooks2023instructpix2pix}, which lacks edits like removal and deformation. 
Expanding training data with additional datasets could improve the quality and diversity of the editing embedding space.
Please refer to the supplementary material for examples of failure cases. 

For future work, \abbrv could be applied to downstream tasks like instruction caption generation, query-based editing pair retrieval, and extensions to video and 3D editing. 
Further improvements include exploring advanced training strategies, such as refined loss functions \cite{zhai2023siglip} or augmented text instructions \cite{fan2023laclip}, and incorporating masks as an extra channel to enhance control over edit regions.



\section{Conclusion}
We proposed \abbrv, a representation-learning approach for image editing that captures how images transform during edits.
Experiments showed that \abbrv achieves state-of-the-art exemplar-based image editing with no computational overhead.
Moreover, we showed that \abbrv serves as a reliable metric for evaluating edit quality and faithfulness to the reference image, aligning closely with human judgment.
Such a metric can accelerate the development of image editing approaches by providing an evaluation metric that aligns better with human judgment compared to existing metrics.

{
    \small
    \bibliographystyle{ieeenat_fullname}
    \bibliography{zbib}
}

\clearpage
\appendix

\section{Quantitative metrics}
Here we explain the commonly-adopted metrics we used in the quantitative evaluation.

\myparagraph{CLIP Score} is calculated as the cosine similarity between the embedding of the output image $I_o$ and the text embedding of the description of the output image $T_o$; the embeddings are from the original CLIP image encoder $\mathcal{F}_\theta$ and CLIP text encoder $\mathcal{G}_\theta$. It can measure how much the output image is aligned with its description. The calculation is as follows:
\begin{equation}
\text{CLIP Score} = \cos\left(\mathcal{F}_\theta(I_o), \mathcal{G}_\theta(T_o)\right),
\end{equation}
where $cos(A, B)=\frac{\mathbf{A} \cdot \mathbf{B}}{\|\mathbf{A}\| \|\mathbf{B}\|}$ ,denoting the cosine similarity.

\myparagraph{CLIP Directional Similarity} calculates the cosine similarity between the difference of the embeddings of the query image $I_q$ and output image $I_o$, against the difference of the embeddings of query image description $T_q$ and output image description $T_o$. The calculation is as follows:
\begin{align}
\text{CLIP Direct. Similarity} = \cos & (\mathcal{F}_\theta(I_q) - \mathcal{F}_\theta(I_o), \notag \\
& \mathcal{G}_\theta(T_q) - \mathcal{G}_\theta(T_o)).
\end{align}
Alternatively, when the text instruction $T$ is available, the calculation becomes:
\begin{equation}
\text{CLIP Direct. Similarity} = \cos\left(\mathcal{F}_\theta(I_q) - \mathcal{F}_\theta(I_o), \mathcal{G}_\theta(T)\right).
\end{equation}
It can measure how much the change of the images matches the change of the text descriptions. Here, the change of the text descriptions (e.g., \emph{A forest in the summer} $\rightarrow$ \emph{A forest in the winter}) implicitly serve as an text instruction (\emph{Change summer to winter}). Note that this is a similar counterpart to our proposed metric \text{\ect}, while \text{\ect} directly measure the change of the images against the change of the text instruction. Please refer to the main paper for the definition for \text{\ect}. 

\myparagraph{$s_{visual}$} is a metric proposed in \cite{nguyen2023visual}, which can be considered as a variant of the CLIP Directional similarity. It calculates the cosine similarity between the difference of the query image embedding and output image embedding, against the difference of the embeddings of the input image $I_i$ and edit image $I_e$. The calculation is as follows:
\begin{equation}
S_{visual} = \cos\left(\mathcal{F}_\theta(I_q) - \mathcal{F}_\theta(I_o), \mathcal{F}_\theta(I_i) - \mathcal{F}_\theta(I_e)\right).
\end{equation}
Note that this is a similar counterpart to our proposed metric \text{\ecec}. Please refer to the main paper for the definition for \text{\ecec}. 

\myparagraph{LPIPS} (Learned Perceptual Image Patch Similarity)\cite{zhang2018lpips} measures the perpetual similarity between the two images. Here we calculate it between the query image $I_q$ and output image $I_o$. Different from the above mentioned metrics, LPIPS serves as a direct evaluation of how much the output image preserves the query image. A lower LPIPS score usually indicate better faithfulness to the query image. However, too low of LPIPS score may suggest insufficient edits.

\section{More qualitative results}
\subsection{Comparison between ours and baselines}
\begin{figure*}[h!]
\centering
\includegraphics[width=1\textwidth, trim={0cm 0cm 0cm 0},clip]{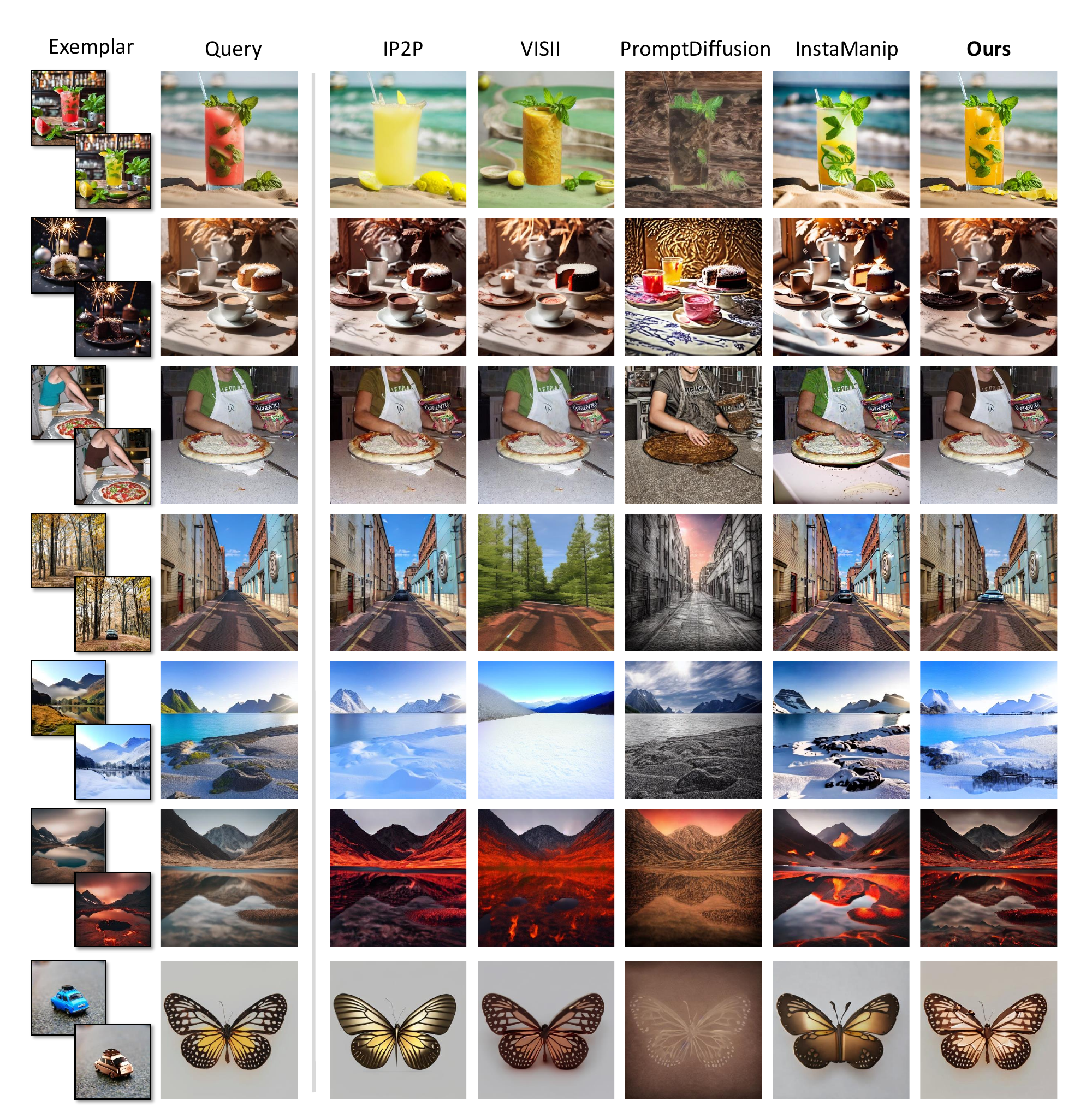}
\caption{More qualitative comparisons between our method and the baselines.}
\label{fig:qualitative_comparison_sup}
\end{figure*}

We show more qualitative comparisons in \cref{fig:qualitative_comparison_sup}. 

\subsection{Transferring edits to multiple test images} 
\begin{figure*}[t!]
\centering
\includegraphics[width=1\textwidth, trim={0cm 8cm 0cm 2},clip]{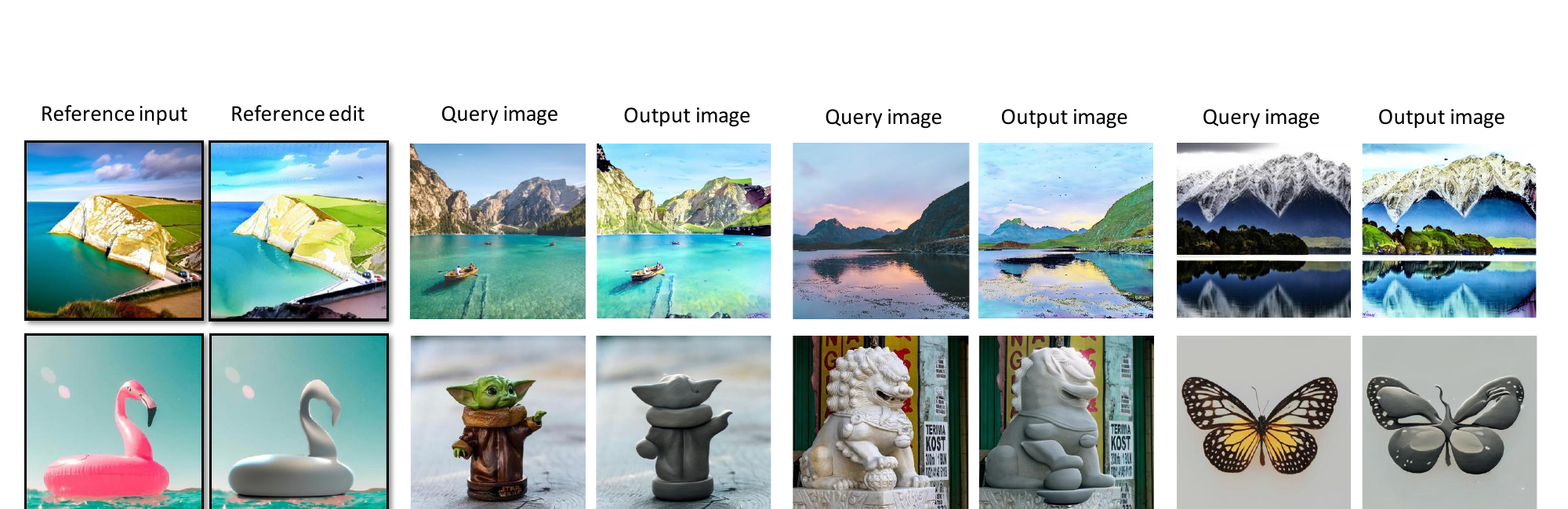}
\caption{Transfer edits from a same exemplar to different test images.}
\label{fig:multitest}
\end{figure*}

We show more visualization of transferring edits from an given exemplar to multiple different test images in \cref{fig:multitest}. We show that the learned embedding of the edits are generalizable to different test images. Note that the test images do not have to be very similar to the exemplars in terms of the low-level structure or style, but rather share high-level similar semantics.

\subsection{Comparison between VIT-B-32 and VIT-L-14}
\begin{figure*}[h!]
\centering
\includegraphics[width=1\textwidth, trim={0cm 6cm 0cm 0},clip]{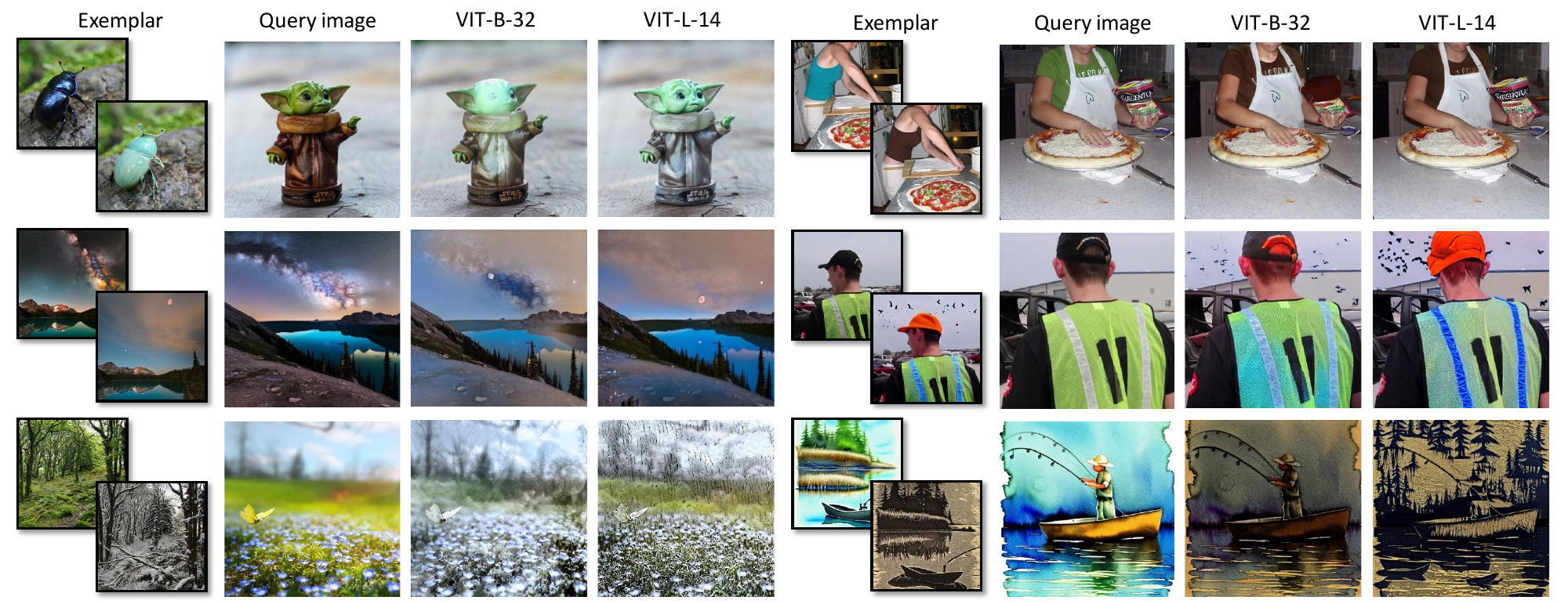}
\caption{Compare the performance between VIT-B-32 and VIT-L-14 as backbone architecture for EditCLIP.}
\label{fig:vit_b_32}
\end{figure*}

We compare the performance between VIT-B-32 and VIT-L-14 as backbone architecture for EditCLIP in \cref{fig:vit_b_32}. We observed that VIT-L-14 achieves a higher quality in most of the cases. While VIT-B-32 can encode the edit from the exemplar, the details of the output image may not be well-preserved (in the first row in \cref{fig:vit_b_32}), or the edit may not be of faithful (in the second row in \cref{fig:vit_b_32}). We conjecture that is because VIT-L-14 is a larger VIT model also with smaller patch sizes, which can capture more visual details compared to VIT-B-32. Therefore, we choose VIT-L-14 as the default backbone for EditCLIP. However, we do found that in some cases when VIT-L-14 struggles to maintain the details when doing global editing applications, VIT-B-32 can well-preserve the original layout details instead (in the third row in \cref{fig:vit_b_32}).

\section{Failure cases}
\begin{figure*}[h!]
\centering
\includegraphics[width=1\textwidth, trim={0cm 6cm 0cm 0},clip]{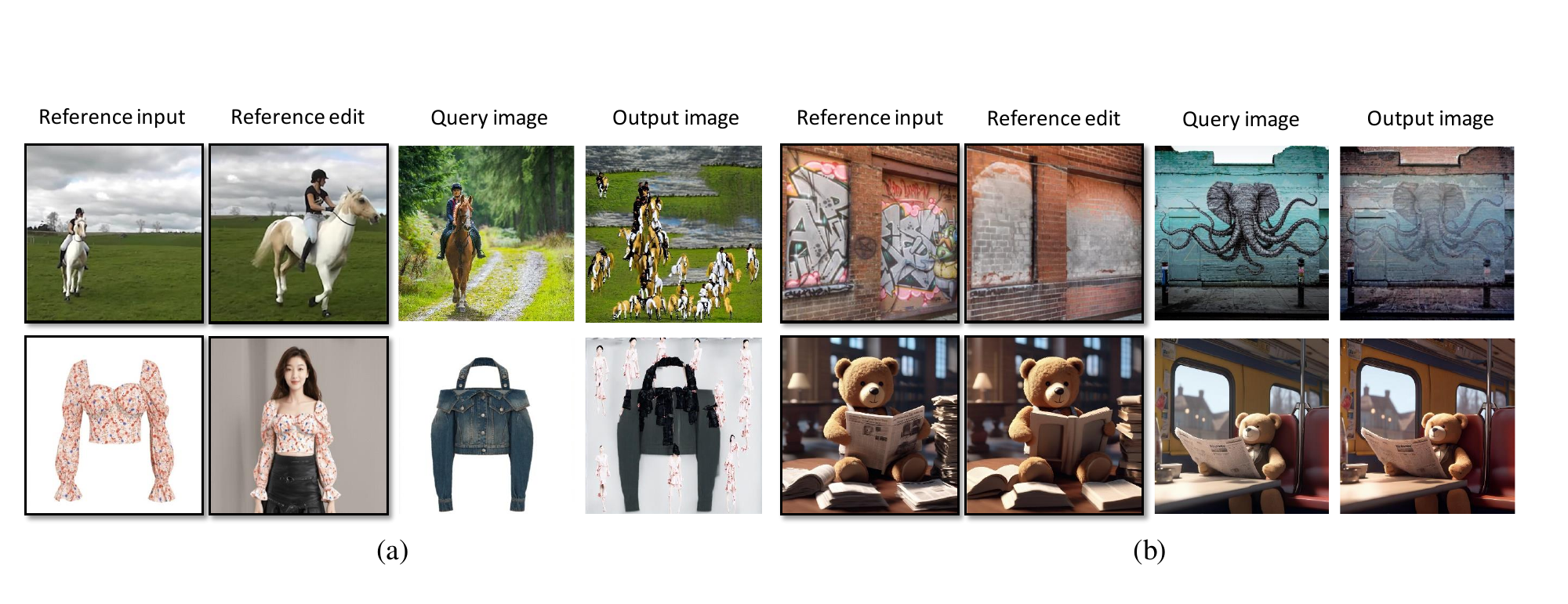}
\caption{Failure cases of our method in exemplar-based image editing.}
\label{fig:failure_cases}
\end{figure*}

We report two types of edits which our method fails to faithfully perform: deformation (in \cref{fig:failure_cases}(a)) and removal (in \cref{fig:failure_cases}(b)). Training datasets which contain these types of edits and potential model architecture designs are needed in order to enable our model for a series of editing applications, such as pose transfer, virtual try-on and removing unwanted objects.

\section{Additional ablation studies}
\subsection{Input loss preservation}

\begin{figure*}[h!]
\centering
\includegraphics[width=1\textwidth, trim={0cm 10cm 0cm 0},clip]{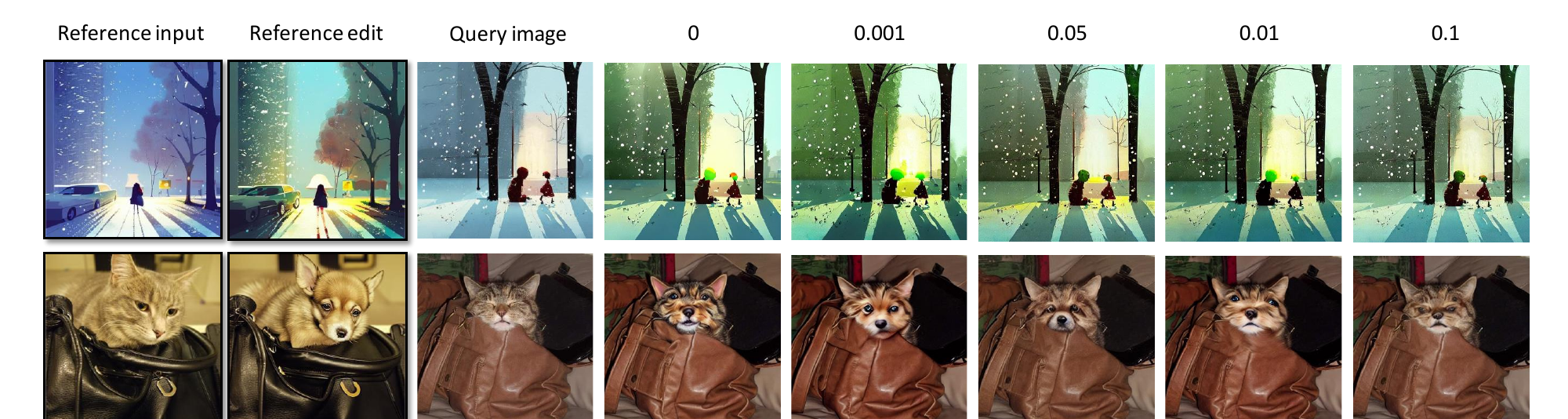}
\caption{The effects of different values of $\lambda_2$.}
\label{fig:lambda_2}
\end{figure*}

We ablate on different values of $\lambda_2$ in \cref{fig:lambda_2}, which control the strength of the input preservation loss against the diffusion denoising loss. When $\lambda_2 = 0$, it means no input preservation loss is applied. Intuitively, larger number of $\lambda_2$ will preserve more input layout, while a smaller one will allow more edits. We balance these two sides and choose 0.05 as the default value for $\lambda_2$.

\subsection{Choice of \abbrv Embedding Layer} 
\begin{figure*}[t!]
\centering
\includegraphics[width=\linewidth, trim={0cm 12cm 0cm 0},clip]{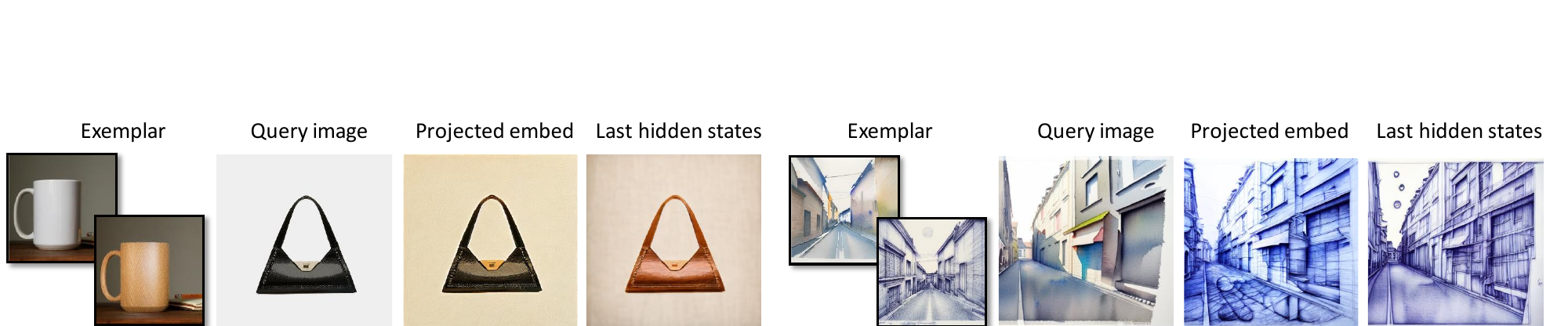}
\caption{Ablation of using the projected embedding after projection layer or hidden states from the last transformer layer from EditCLIP for the embedding.}
\label{fig:ablation_last_hidden}
\end{figure*}

Different from the common practice \cite{Ramesh2022unclip,ye2023ip-adapter} that uses the projected embedding from CLIP as the image condition, we found that using hidden states from the last transformer layer before going to the CLIP projection layer is more effective to transfer the edit while preserving the input layout. 
\Cref{fig:ablation_last_hidden} that in our task, we found 
We conjecture that it is because last hidden states contain more tokens, which encode more visual details and hence have higher capacity in general.

\subsection{Guidance scale}
As it is done in \cite{brooks2023instructpix2pix}, our denoising UNet for exemplar-based editing is also conditioned on both the VAE input image $\mathcal{E}(I_i)$ and edit embedding $E$. Therefore, during inference, we could apply two separate guidance scales similar to \cite{brooks2023instructpix2pix}, where edit guidance scale $s_E$ controls how the output image follows the edits, and image guidance scale $s_I$ controls how the output image resembles the input image. 

The modified score estimate $\tilde\epsilon_{\theta}$ is as follows:  
\begin{align}
\tilde\epsilon_{\theta}(x_t, t, \mathcal{E}(I_i), E) &= \epsilon_{\theta}(x_t, t, \varnothing, \varnothing) \notag \\
&+ s_I \left( \epsilon_{\theta}(x_t, t, \mathcal{E}(I_i), \varnothing) - \epsilon_{\theta}(x_t, t, \varnothing, \varnothing)\right) \notag \\
&+ s_E \left( \epsilon_{\theta}(x_t, t, \mathcal{E}(I_i), E) - \epsilon_{\theta}(x_t, t, \mathcal{E}(I_i), \varnothing)\right)
\end{align}

\begin{figure*}[t!]
\centering
\includegraphics[width=\linewidth, trim={0cm 4cm 0cm 0},clip]{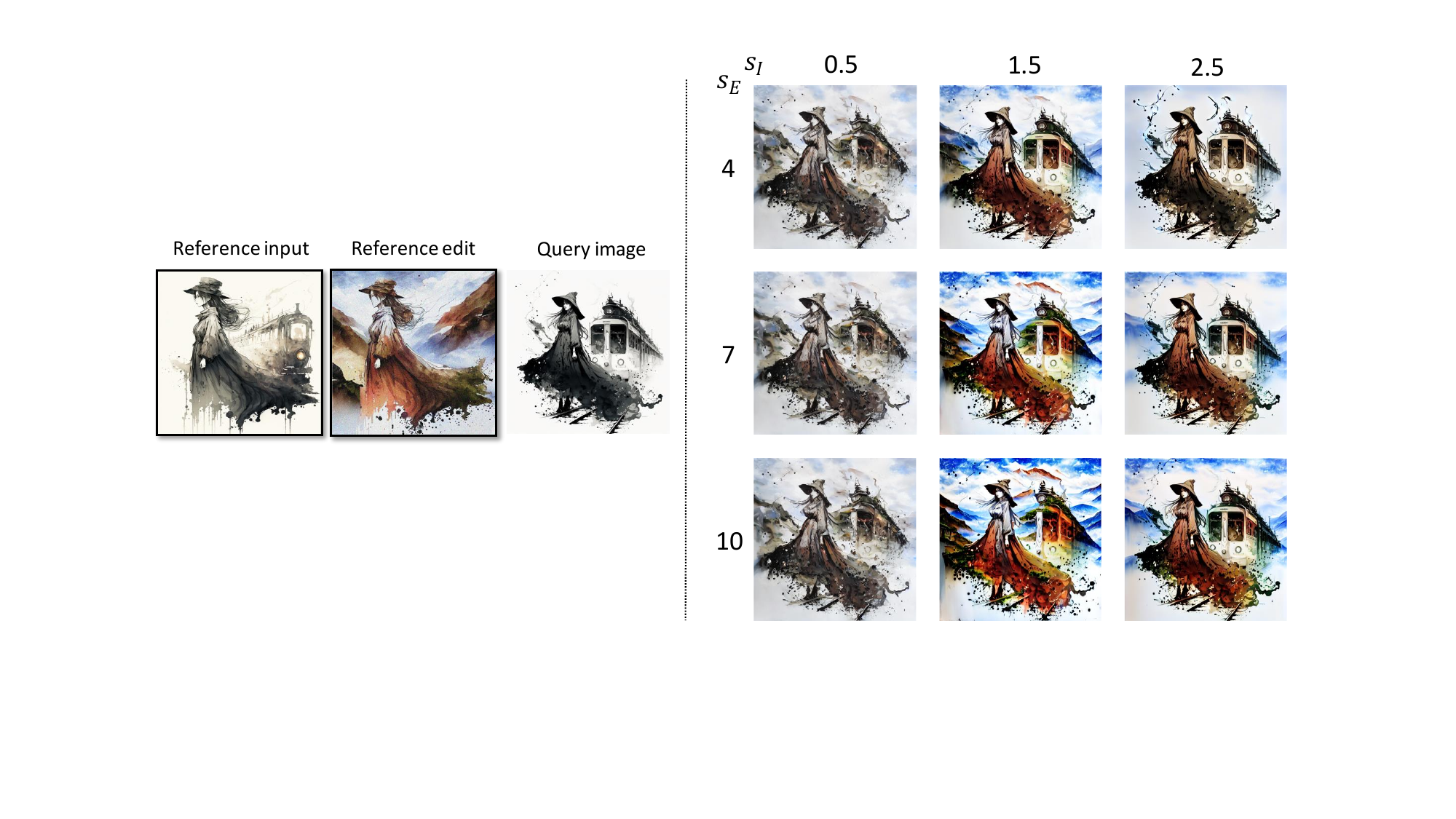}
\caption{Ablation of using different values for guidance scales $s_E$ and $s_I$.}
\label{fig:ablation_guidance_scale}
\end{figure*}

We show the ablation of the guidance scales in \cref{fig:ablation_guidance_scale}. In general, as $s_E$ increases, the output images will have stronger editing effects; while when $s_I$ increases, the output images will follow more the input image. By default, we set $s_E=7$ and $s_I=1.5$, which is the suggested practice in \cite{brooks2023instructpix2pix}. However, users can tune these hyperparameters to obtain desired results.

\section{Benchmark statistics}
We adapt the TOP-Bench dataset \cite{zhao2024instructbrush} for exemplar-based image editing and we denote it as \emph{TOP-Bench-X}.  
TOP-Bench consists of different types of edits, where each type includes a set of training and test pairs.  
We use the training set to form exemplar pairs, denoted as $[I_i, I_e]$, while the test set provides the corresponding query image $I_q$.
This results in a total of 1277 samples, comprising 257 unique exemplars and 124 unique queries. Edit types contain between 32 and 60 samples. In addition to query-exemplar pairing, we perform multiple seeds per method, for the metric evaluation we include all the seeds. 

\begin{figure*}
    \centering
    \includegraphics[width=1.0\linewidth]{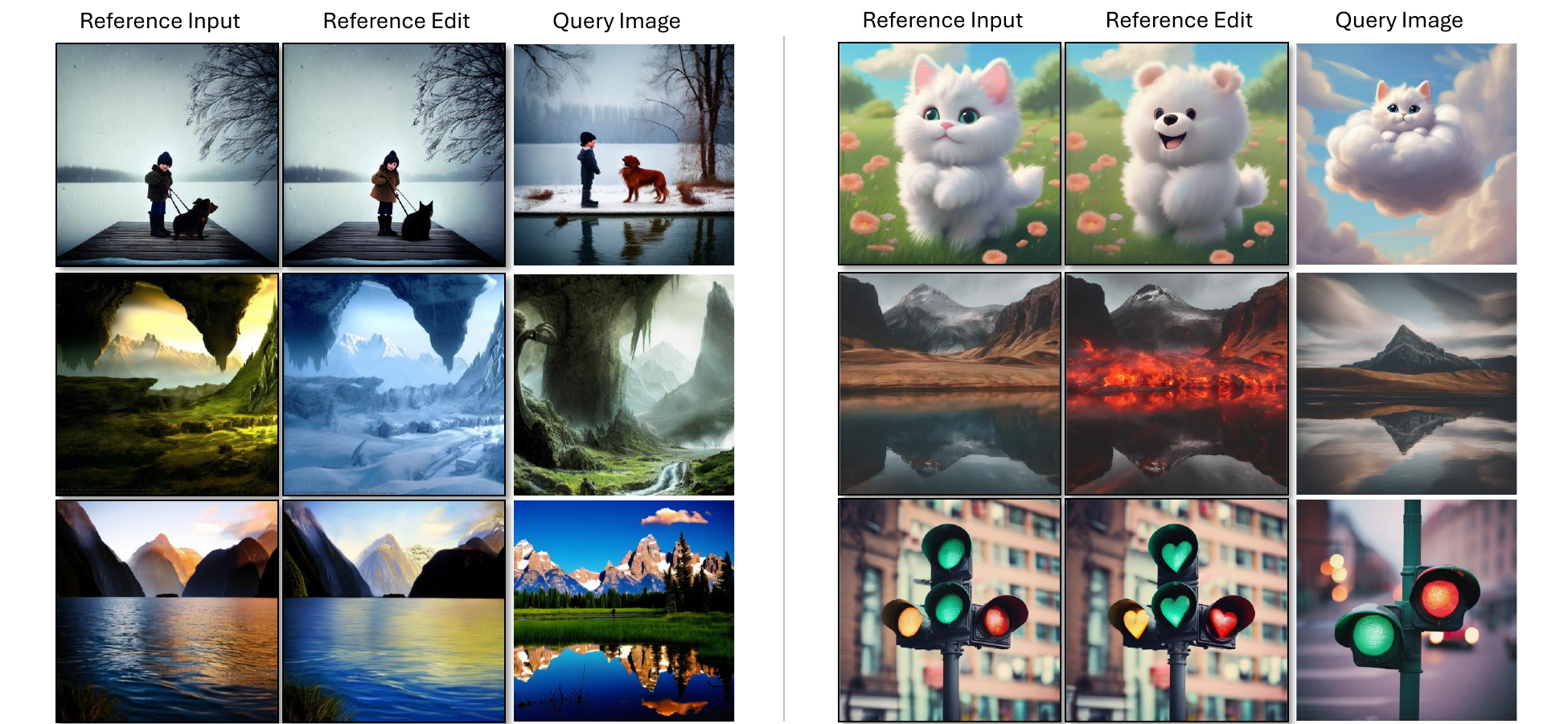}
    \caption{Additional visualization of exemplars present in the \emph{TOP-Bench-X} variant. }
    \label{fig:exemplar_example}
\end{figure*}

We visualize additional exemplar pairs with queries from the benchmark on \cref{fig:exemplar_example}, where we can see different types of edits present in the benchmark. 

\section{User study}

\begin{figure*}
    \centering
    \includegraphics[width=1.0\linewidth]{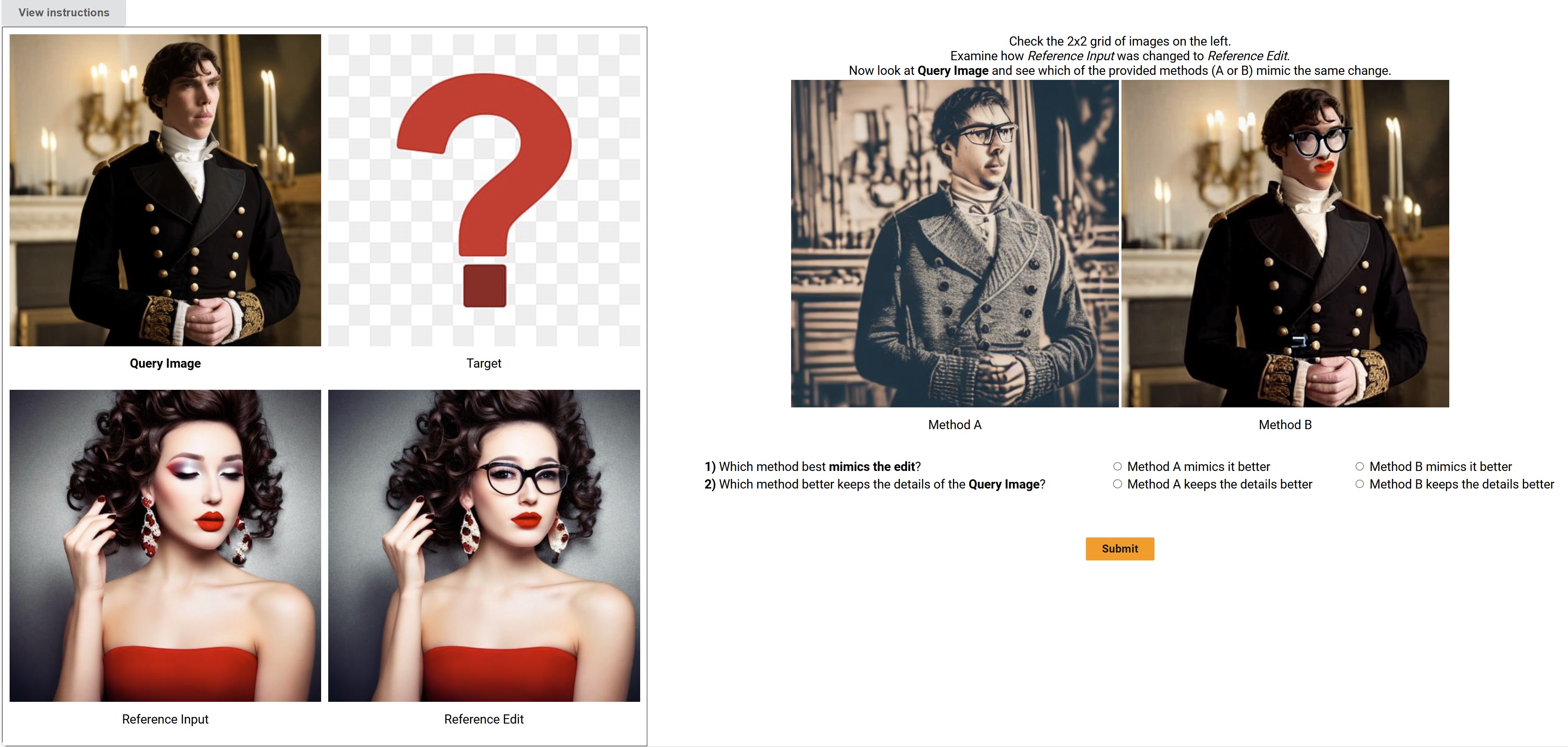}
    \caption{Single example of the 2AFC user study. Participants see the Query and Exemplar pairs on the left and two potential edits on the right. They are asked to select which method best mimics the edit and which better preserves the Query image details. }
    \label{fig:user_study}
\end{figure*}

The user study was conducted on Amazon MTurk with two alternative forced-choice (2AFC) layout as seen on \cref{fig:user_study}. We use only participants with Master Qualification on the platform. There were a total of 53 unique participants, with the average time of each sample taking 40 seconds, and the average user did 89 samples with a total of 4712 comparisons. We randomly select 2 seeds (out of 5 seeds) for each inference.

\end{document}